\documentclass{article}

\usepackage{PRIMEarxiv}

\usepackage[utf8]{inputenc} 
\usepackage[T1]{fontenc}    
\usepackage{hyperref}       
\usepackage{url}            
\usepackage{booktabs}       
\usepackage{amsfonts}       
\usepackage{nicefrac}       
\usepackage{microtype}      
\usepackage{lipsum}
\usepackage{fancyhdr}       
\usepackage{graphicx}       
\graphicspath{{media/}}     

\title{Intelligent Conversational Android ERICA Applied to Attentive Listening and Job Interview
}

\author{
  Tatsuya Kawahara, Koji Inoue, Divesh Lala \\
  Graduate School of Informatics \\
  Kyoto University \\
  Kyoto, Japan\\
  \texttt{\{kawahara, inoue, lala\}@sap.ist.i.kyoto-u.ac.jp} \\
}

\begin{document}
\maketitle

\begin{abstract}
Following the success of spoken dialogue systems (SDS) in smartphone assistants and smart speakers, a number of communicative robots are developed and commercialized.
Compared with the conventional SDSs designed as a human-machine interface, interaction with robots is expected to be in a closer manner to talking to a human because of the anthropomorphism and physical presence.
The goal or task of dialogue may not be information retrieval, but the conversation itself. In order to realize human-level ``long and deep'' conversation, we have developed an intelligent conversational android ERICA.
We set up several social interaction tasks for ERICA, including attentive listening, job interview, and speed dating.
To allow for spontaneous, incremental multiple utterances, a robust turn-taking model is implemented based on TRP (transition-relevance place) prediction, and a variety of backchannels are generated based on time frame-wise prediction instead of IPU-based prediction.
We have realized an open-domain attentive listening system with partial repeats and elaborating questions on focus words as well as assessment responses.
It has been evaluated with 40 senior people, engaged in conversation of 5-7 minutes without a conversation breakdown.
It was also compared against the WOZ setting.
We have also realized a job interview system with a set of base questions followed by dynamic generation of elaborating questions.
It has also been evaluated with student subjects, showing promising results.
\end{abstract}


\section{Introduction}

Spoken dialogue systems (SDS) have been deployed in many gadgets such as smartphone assistants and smart speakers in the past ten years.
Millions of users now use them daily. In these systems, dialogue is regarded as a means to complete some tasks by the machine.
The typical tasks include command and control of the machine or application software (e.g., music player and alarm) and simple information retrieval (e.g., weather and routing).
The task goals are objective and shared by the system and users.
Thus, the dialogue should be completed as soon as possible.
There is a big gap from the human-human dialogue, in which task goals are not definite and the duration is flexible.

Conversational robots are another application of the spoken dialogue systems, but they should be designed in a different principle from ``smart'' devices mentioned above.
We do not need a robot to ask for weather information and alarm.
Instead, the conversational robots are expected to simply talk to or listen to humans, as humans do.
An ultimate goal is to realize human-level dialogue, which is engaging and pleasant.
Apart from the quality of dialogue, there is a significant difference in the style of the dialogue.
In the current human-machine interface, a user is assumed to utter one sentence per one turn, which corresponds to a command or a query, to which the system will respond.
On the other hand, in natural human dialogue, a participant utters many sentences per one turn, during which the counterpart makes backchannels.
The difference can be analogous to half-duplex versus full-duplex in the communication channel.
It is weird to nod or backchannel to a machine, but a human-looking android would help the realization of human-level dialogue.

With these backgrounds, we are conducting a project to develop an autonomous android ERICA~\cite{dylan2016roman,inoue2016sigdial,kawahara2018iwsds}, who behaves and interacts just like a human, including facial look and expression, gaze and gesture, and spoken dialogue.
An ultimate criterion is to pass a Total Turing Test, that is to convince people that ERICA’s interaction is comparable to a human, or it is indistinguishable from a remote-operated android.
We hope that through this project we can make clear what is missing or critical in natural interaction.
ERICA is expected to replace some social roles currently done by a human, or used for conversation skill training in a realistic setting.

In this article, the major challenges and key technologies in this autonomous android are described.
In particular, the systems developed for attentive listening and job interview are explained with evaluations.

\section{Android ERICA}

\begin{figure}[t]
  \begin{center}
  \includegraphics[width=90mm]{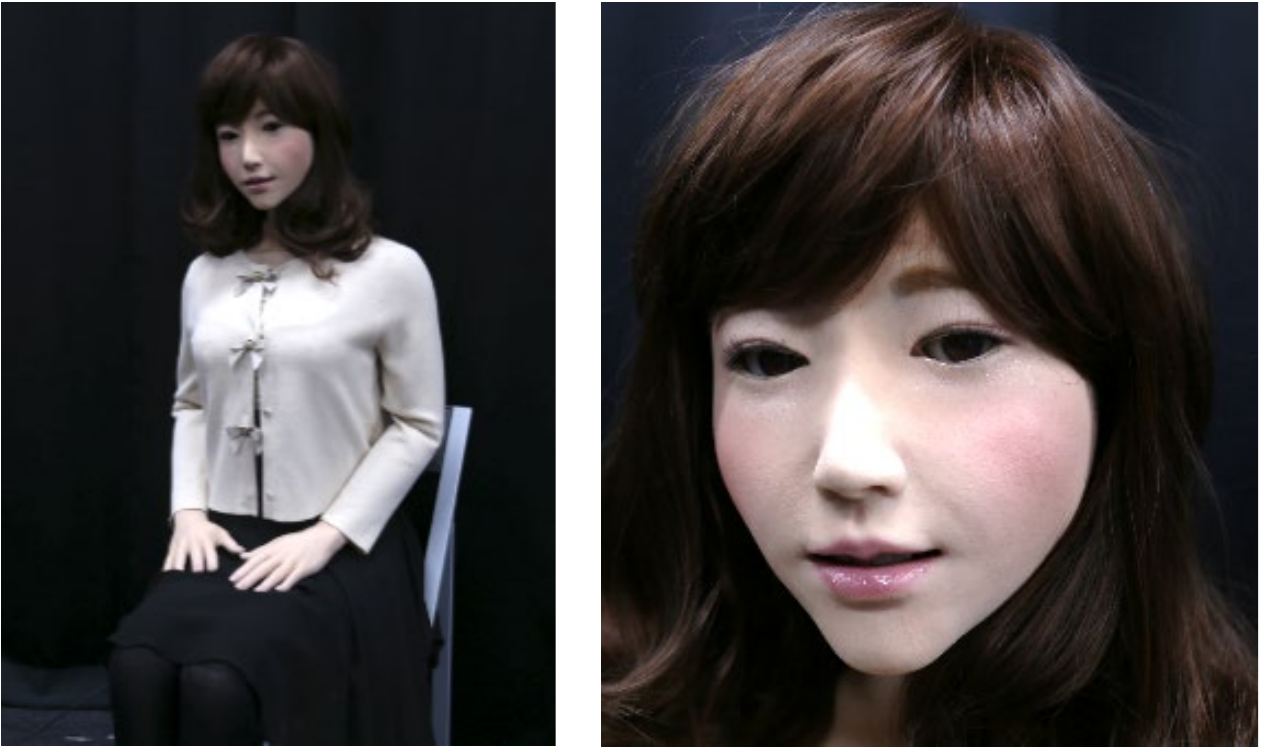}
  \end{center}
  \caption{Android ERICA}
  \label{fig:erica}
\end{figure}

Figure~\ref{fig:erica} shows a snapshot of ERICA.
She does not move around, but generates elaborate facial expressions and movements of heads, eyes, and body attitudes while sitting on a chair, so that she can behave and interacts exactly like a human.
We have also developed ASR (automatic speech recognition) and TTS (text-to-speech) systems as well as the human tracking system dedicated to ERICA.

\subsection{Social interaction tasks}

We have explored a number of social roles suited to ERICA that can take advantage of human-like presence and would realize human-like dialogue.
A majority of tasks conducted by the current SDSs such as information services are not adequate; they are better suited to smartphones and smart speakers.
While most conventional robots are engaged in physical tasks such as moving objects or delivering goods, recently many kinds of communicative robots are designed and introduced in public spaces.
They serve as an attendant~\cite{fujie2009conversation} or a receptionist~\cite{bohus2016models}.
They are effective for attracting people, but the resulting interaction is usually very shallow and short such as a predefined script and simple guidance.
In recent years, chatting systems are also developed intensively, but the dialogue is also shallow and not so engaging. In contrast, interaction with ERICA should leverage physical presence and involve face-to-face communication.
Therefore, we design ``social interaction'' tasks in which human-like presence matters and long deep interaction is exchanged. Here, dialogue itself is a task, and the goal of dialogue may be mutual understanding or appealing. We assign a realistic social role to ERICA, so matched users are seriously engaged beyond chatting. Specifically, we set up the following four tasks.
They are compared in Table~\ref{table:task_for_erica}.

\subsubsection{Attentive listening}

In this task, ERICA mostly listens to senior people talking about topics such as memorable travels and recent activities~\cite{lala2017}.
Attentive listening is being recognized as effective for maintaining the communication ability of senior people, and many communicative robots are designed for this task.
The role of ERICA is to encourage users to speak for long.
In this sense, attentive listening is similar to counseling~\cite{devault2014simsensei}.

\subsubsection{Job interview (practice)}

While dialogue systems have been investigated for casual interviews~\cite{kobori2016small}, a job interview is very important for both applicants, typically students, and companies hiring them.
Each side makes a lot of preparations including rehearsal.
In this setting, ERICA plays the role of interviewer by asking questions.
She provides a realistic simulation, and is expected to replace a human interviewer in the future.

\subsubsection{Speed dating (practice)}

Speed dating is widely held for giving an opportunity for people to find a partner.
In this setting, two persons meet for the first time and talk freely to introduce themselves, and see if the counterpart can be a good match.
There was a study~\cite{ranganath2009s} that analyzed a corpus of speed dating.
In our setting, ERICA plays the role of the female participant by talking about topics such as hobbies and favorite foods.
She provides a realistic simulation and gives proper feedbacks according to the dialogue.

\subsubsection{Lab guide}

A robot is often used for introducing laboratories and museums.
It is expected to attract people, and needs to keep their engagement by providing proper interaction.

\begin{table}[t]
  \caption{Dialogue tasks designed for ERICA}
  \label{table:task_for_erica}
  \begin{center}
  \begin{tabular}{lcccc}
  \hline
  & attentive listening & job interview & speed dating & lab guide \\
  \hline
  role of system & listen & ask & all & talk \\
  dialogue initiative & user & system & mixed & system \\
  main speaker & user & user & both & system \\
  main listener & system & system & both & user \\
  turn-taking & few & explicit & complicate & explicit \\
  \hline
  \end{tabular}
  \end{center}
\end{table}

\begin{figure}[t]
  \begin{center}
  \includegraphics[width=150mm]{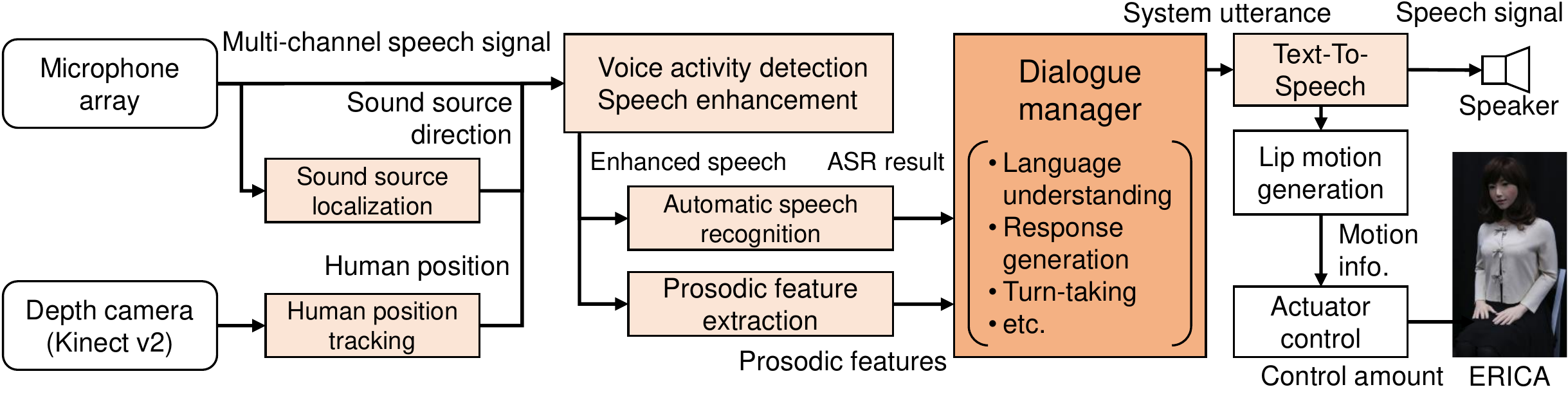}
  \end{center}
  \caption{System configuration for android ERICA}
  \label{fig:system}
\end{figure}

\subsection{System configuration} \label{sec:system}

Figure~\ref{fig:system} depicts the whole system configuration for android ERICA.
The input devices are a 16-channel microphone array and a depth camera (Kinect v2).
These sensors are placed within the interaction environment but not physically within or on the android robot.
The sensors do not require any human contact or pre-dialogue calibration, so users can start and engage in dialogue naturally without awareness of any sensors (as opposed to using hand microphones, for example).
We use the sensors for sound source localization, voice activity detection, and speech enhancement to obtain the segmented speech of the target user~\cite{carlos2016iros}.
The enhanced speech is then fed into the ASR (automatic speech recognition) module which uses an acoustic-to-subword end-to-end neural network model.
Simultaneously, prosodic information such as fundamental frequency (F0) and power is extracted from the enhanced speech~\cite{carlos2016iros}.
The ASR result is then fed into the dialogue manager to generate a system response which will be then played by a TTS (text-to-speech) engine designed for ERICA\footnote{\url{https://voicetext.jp/news/product/151023/}}.
The TTS engine is also enabled to play non-linguistic utterances such as backchannels, fillers, and laughing.
Lip and head movements are controlled based on the generated TTS speech~\cite{carlos2012iros,sakai2015roman}.


\section{Technologies for human-level conversations}

For human-level conversation, linguistic processing and understanding is necessary but non-linguistic behaviors are similarly important. 
We briefly explain several of these components for ERICA.

\subsection{Backchannel generation}

Backchannels are short utterances uttered by listeners such as ``{\it yeah}'' in English and ``{\it un}'' in Japanese.
Listeners utter backchannels toward speakers to stimulate further talk and also to express the listener's attention and interest in the conversation.
In order to generate backchannels, systems need to predict the timing of backchannels, which has been tackled by many works using prosodic features~\cite{ward2000prosodic,truong2010rule}.
While Existing backchannel generation systems make a prediction of backchanneling after the end of user utterances segmented by IPUs (inter-pausal units), we implement frame-wise continuous prediction of backchannel timing with a logistic regression that predicts if the system should utter a backchannel within the next 500 milliseconds~\cite{lala2017}.
We also proposed a prediction model of backchannel form (type) based on both prosodic and linguistic features~\cite{kawahara2016prediction}.

\subsection{Flexible turn-taking using TRP}

Flexible and robust turn-taking is a required function for natural communication between the system and a user.
End-of-turn prediction has been widely studied using linguistic and prosodic features extracted from the user's utterance~\cite{skantze2017sigdial,masumura2017interspeech}.
Existing models were trained with actual turn-taking behaviors in dialogue corpora, but the turn-taking decision is sometimes arbitrary so model training can become unstable.
Therefore, we take into account the concept of TRPs (transition-relevance places)~\cite{sacks1974TRP} where the current turn could be completed, and are independent of the actual end-of-turn instances.
We manually annotated these TRP instances in our human-robot dialogue corpus~\cite{hara2019interspeech}.
We then proposed a two-step turn-taking prediction model where the system first detects the TRP, and if detected, predicts if the system actually takes the turn.
By decomposing complex turn-taking prediction into TRP detection and subsequent turn prediction, the accuracy of turn-taking was improved.

From the perspective of live spoken dialogue systems, another required function for the turn-taking module is to predict how long the system waits until it actually takes the turn.
A simple approach is to set a fixed silence threshold time, but it may cause false cut-in (interruption) or unnaturally long silences.
Our system integrates the above-mentioned turn-taking prediction model using TRP labels with an FSTTM (Finite-State Turn-Taking Machine)~\cite{Raux2009,Lala2018} to determine the silence time the system will wait for.
It can quickly take the turn with a high probability of being end-of-turn, while it can also wait longer if the user says utterances such as fillers or hesitations.

\subsection{Engagement recognition}

Engagement is defined as how much a user is interested in the current dialogue, and keeping users engaged is an important factor that leads to ``long and deep'' dialogue~\cite{oertel2020engagement}.
To this end, engagement recognition has been widely studied by investigating multi-modal user behaviors~\cite{dhall2019emotiw}.
We proposed an engagement recognition model based on listener behaviors such as backchannels, laughs, head nods, and eye contact~\cite{inoue2018engagement}.
We automatically detected these listener behaviors to implement a real-time engagement recognition system.
This system was applied to the laboratory guide dialogue task with ERICA to control the android robot's behaviors according to the level of user engagement~\cite{inoue2019iwsds}.

\begin{figure}[t]
  \begin{center}
  \includegraphics[width=160mm]{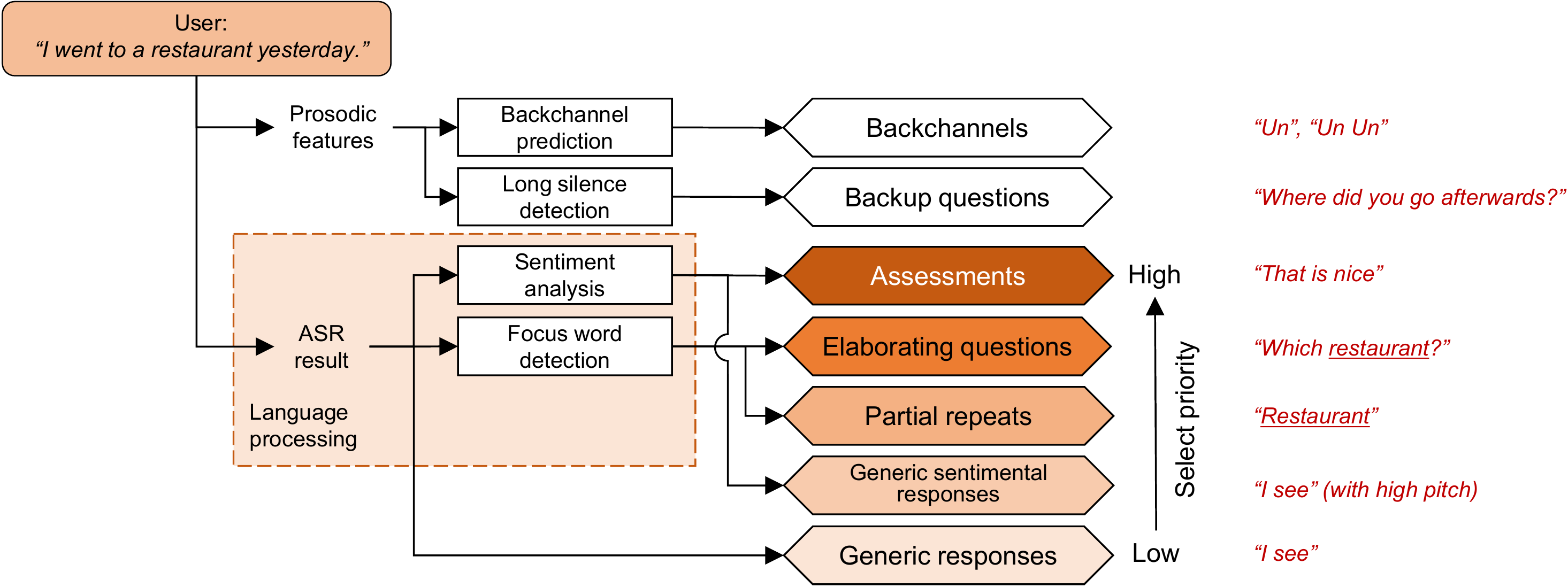}
  \end{center}
  \caption{List of listener responses in our attentive listening system (Examples of generated responses are shown in right side in this figure. Underline means the focus focus word.)}
  \label{fig:response_list}
\end{figure}

\begin{figure}[t]
  \begin{center}
  \includegraphics[width=80mm]{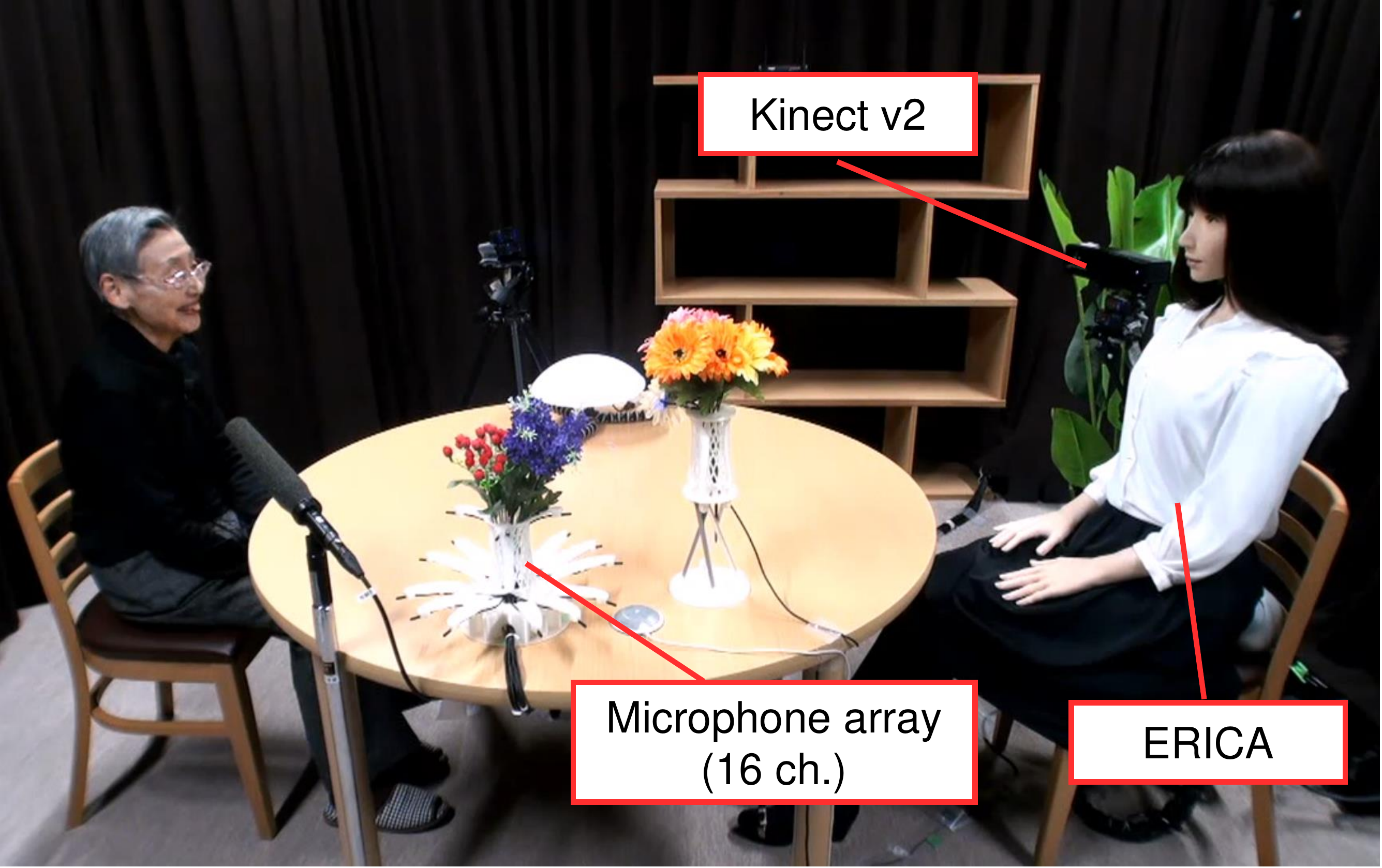}
  \end{center}
  \caption{Snapshot of dialogue experiment for our attentive listening system with ERICA}
  \label{fig:experiment}
\end{figure}

\section{Attentive Listening System}

We implemented an attentive listening system with ERICA~\cite{lala2017,inoue2020sigdial}.
The aforementioned backchannel generation model is a core component in this system.
Besides backchannels, the system generates a variety of listener responses: partial repeats, elaborating questions, assessments, generic sentimental responses, and generic responses, as depicted in Figure~\ref{fig:response_list}.
Partial repeats and elaborating questions are generated based on the {\it focus word} of user utterances.
Assessments and generic sentimental responses are based on the sentiment (positive or negative) of user utterances.

We have conducted dialogue experiments with a total of 40 senior people and confirmed that they could engage with ERICA for 5-7 minutes without a conversation breakdown.
Figure~\ref{fig:experiment} shows the snapshot of this experiment.
We also manually evaluated each system response and found that about 60\% of the system responses were acknowledged as appropriate, which means the other 40\% responses can be improved.
The novelty of our experiment is to compare our system with a human listener implemented in a WOZ (Wizard-of-OZ) setting with a hidden human operator tele-operating ERICA~\cite{inoue2020sigdial}.
From the subjective evaluation, our system achieved comparable scores against the WOZ setting in basic skills of attentive listening such as {\it encouragement to talk}, {\it focused on the talk}, and {\it actively listening}.
On the other hand, there is still a gap between our system and human listeners for more sophisticated skills such as {\it dialogue understanding}, {\it showing interest}, and {\it empathy towards the user}.
From this experiment we could identify the type of skills that differed between our system and human listeners.
In future work, we will improve the response generation modules to increase the ratio of appropriate responses and close the gap with human listeners.


\section{Job Interview System}

We also implemented a job interview system where ERICA plays the role of an interviewer~\cite{inoue2020icmi}.
Existing job interview systems give the same pre-defined questions to any interviewees, so interviews tend to be tedious and far from real human-human interviews.
To make interviews more realistic, our system generates follow-up questions dynamically based on initial responses from the interviewee with two different approaches.
The first approach is based on assessing the quality of the interviewee's utterance using a checklist of items that ``{\it should be mentioned}'' in the job interview.
For example, when the interviewee could not mention what he or she wants to contribute to the company, the system generates a follow-up question such as ``{\it Which part of our company do you want to contribute to?}''. 
The second approach is based on keyword extraction from the interviewee's response.
For example, when the interviewee said ``{\it I have an experience on machine learning.}'', then the system extracts a keyword as ``{\it machine learning}'' and then generates a follow-up question such as ``{\it Could you explain more about machine learning?}''

\begin{figure}[t]
  \begin{center}
  \includegraphics[width=120mm]{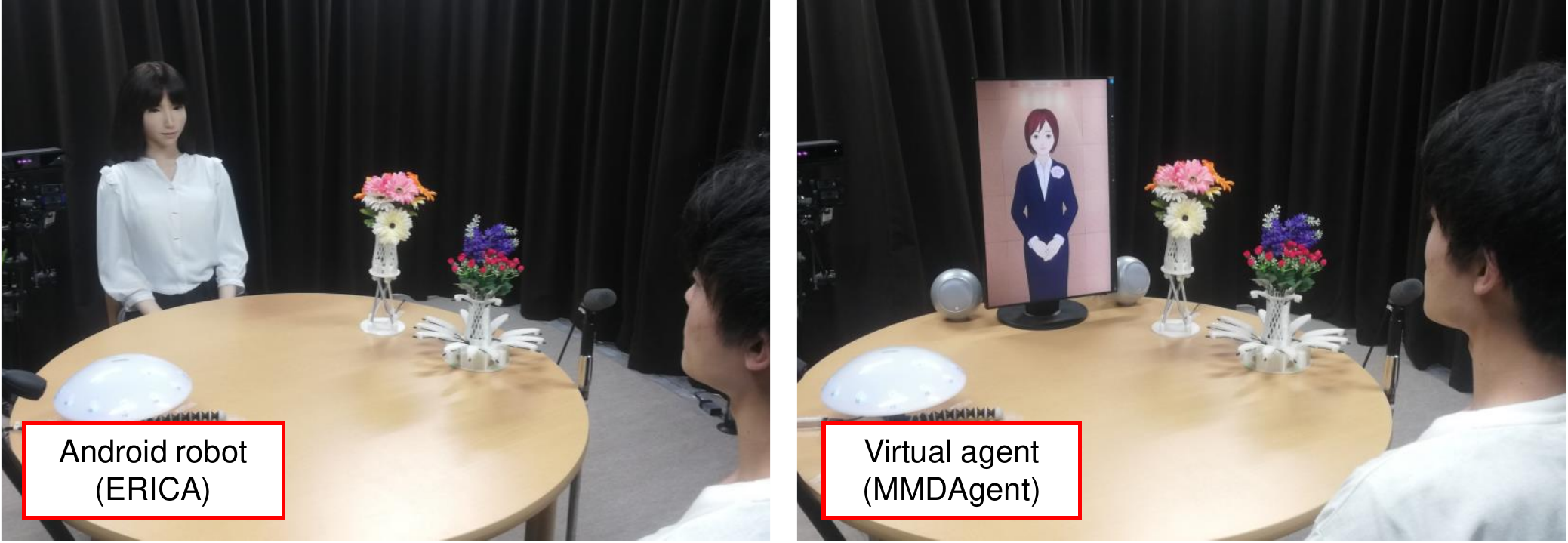}
  \end{center}
  \caption{Difference of appearance between ERICA and virtual agent in our job interview dialogue experiment}
  \label{fig:mmd}
\end{figure}

We conducted a dialogue experiment with university students to compare our system with a baseline system that asked only pre-defined basic questions.
We found that our system was significantly better than the baseline system in the quality of the questions.
It was also found that the perceived presence of the android interviewer was enhanced by the follow-up questions~\cite{inoue2020icmi}.
Another research question in this job interview scenario is how much the interviewer's appearance affects the above-reported experimental result, so we also conducted the same experiment with a virtual agent, as shown in Figure~\ref{fig:mmd}.
As a result, we confirmed a similar result for the follow-up questions even with the virtual agent interviewer, but the presence of the virtual agent interviewer was not enhanced.
The presence of the interviewer is an important factor in creating an interview system with a realistically tense atmosphere.
Therefore, it is more effective to implement the follow-up questions with android ERICA as the job interviewer than a virtual agent.

\section{Conclusions}

Because of the pervasiveness of SNS as a communication channels and then COVID-19, importance of face-to-face communication is questioned or emphasized.
The android ERICA provides a testbed for investigating what are important factors for the quality communications.
The android is expected to replace some social roles under and after the COVID-19 era.

\section*{Acknowledgments}
This work was supported by JST ERATO Grant number JPMJER1401 and JSPS KAKENHI Grant number JP19H05691.

\bibliographystyle{unsrt}  
\bibliography{references}

\end{document}